\documentclass[runningheads]{llncs}

\usepackage{eccv}

\usepackage{eccvabbrv}

\usepackage{graphicx}
\usepackage{booktabs}

\usepackage[accsupp]{axessibility}  

\usepackage{hyperref}
\usepackage{algorithm,algpseudocode}
\usepackage{orcidlink}

\begin{document}

\title{Closer to Ground Truth: Realistic Shape and Appearance Labeled Data Generation for Unsupervised Underwater Image Segmentation} 

\titlerunning{Realistic Shape and Appearance Labeled Underwater Fish Data Generation}

\author{Andrei Jelea\inst{1,2}\orcidlink{0009-0002-3809-9387} \and 
Ahmed Nabil Belbachir \inst{1}\orcidlink{0000-0001-9233-3723} \and
Marius Leordeanu\inst{1,2,3}\orcidlink{0000-0001-8479-8758}}

\authorrunning{A. Jelea, A.N. Belbachir and M. Leordeanu}

\institute{NORCE Norwegian Research Center AS \and
National University of Science and Technology "Politehnica" Bucharest \and
"Simion Stoilow" Institute of Mathematics of the Romanian Academy}

\maketitle

\nocite{92} 
\begin{abstract}
  Solving fish segmentation in underwater videos, a real-world problem of great practical value in marine and aquaculture industry, is a challenging task due to the difficulty of the filming environment, poor visibility and limited existing annotated underwater fish data. In order to overcome these obstacles, we introduce a novel two stage unsupervised segmentation approach that requires no human annotations and combines artificially created and real images. Our method generates challenging synthetic training data, by placing virtual fish in real-world underwater habitats, after performing fish transformations such as Thin Plate Spline shape warping and color Histogram Matching, which realistically integrate synthetic fish into the backgrounds, making the generated images increasingly closer to the real world data with every stage of our approach. While we validate our unsupervised method on the popular DeepFish dataset, obtaining a performance close to a fully-supervised SoTA model, we further show its effectiveness on the specific case of salmon segmentation in underwater videos, for which we introduce DeepSalmon, the largest dataset of its kind in the literature (30 GB). Moreover, on both datasets we prove the capability of our approach to boost the performance of the fully-supervised SoTA model.
\end{abstract}

\section{Introduction}
Fish segmentation in underwater videos plays an essential role in marine and aquaculture applications such as fish body measurement, counting, behaviour analysis and fish motion trajectory estimation. The difficulties of this problem arise from the limited underwater optical view and lack of image clarity, which come from the nature of this kind of data, where illumination is often so poor that fish, underwater shadows and other shapes in the background can be easily confused even by the human eye.
Since approaches that require a good image clarity, such as those using optical flow~\cite{1,2}, do not work in environments of such limited visibility, the majority of existing fish segmentation models rely on heavy supervised training \cite{5,6,7,8}. 
Therefore, since there are very few available annotated fish datasets in the literature, it becomes necessary to create automated labelling methods suitable for the underwater scenario.

\begin{figure}[t]
\begin{center}
   \includegraphics[width=0.97\linewidth]{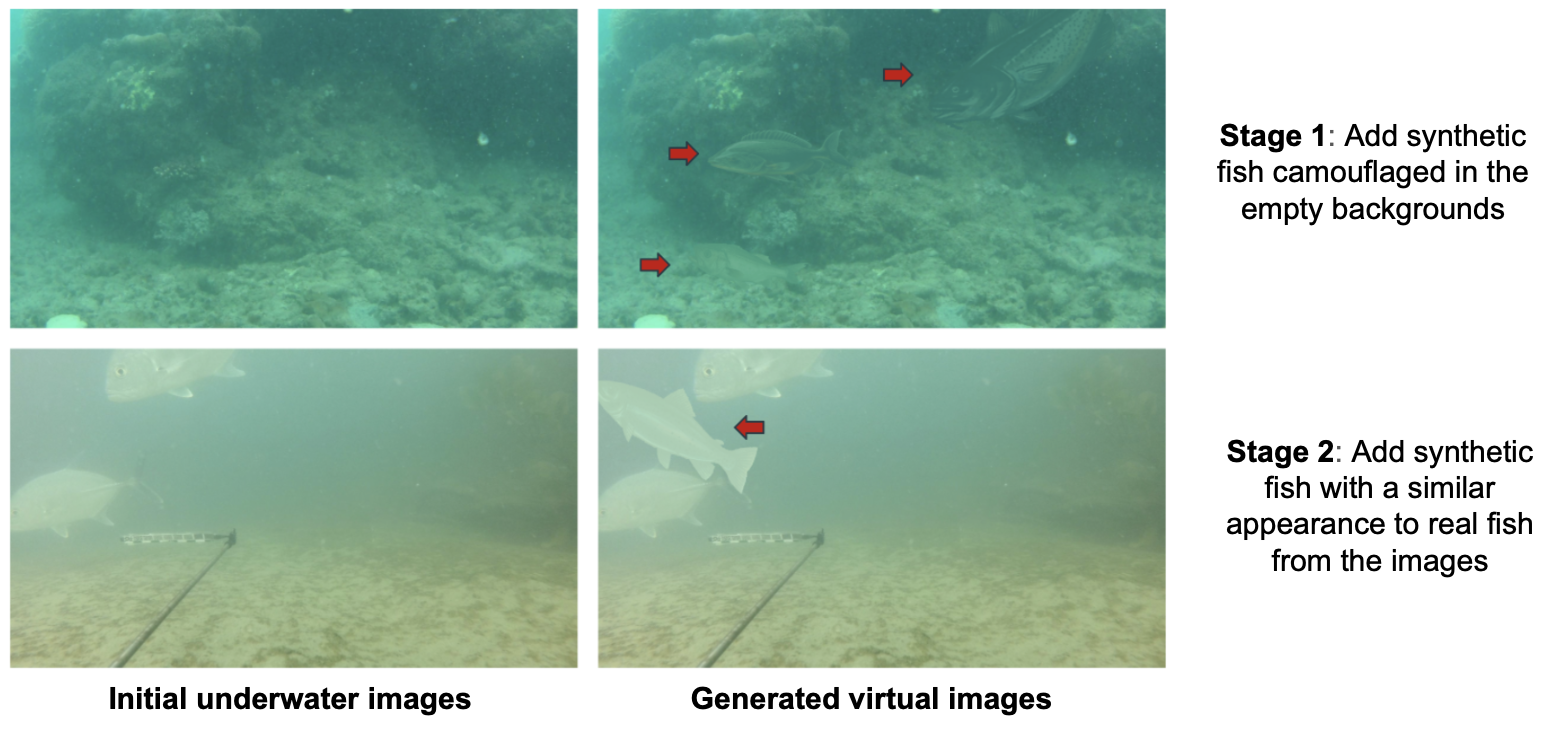}
\end{center}
   \caption{Examples of \textcolor{purple}{synthetic fish} placed into images from DeepFish dataset at both stages of our approach. Note how virtual fish naturally fits in the underwater images due to the data transformations performed. }
\label{fig:add}
\end{figure}

{\bf Our contributions: } We introduce a novel labeled data generation approach that addresses both limitations of poor visibility and lack of annotated data, encountered when solving the task of underwater fish segmentation. Leveraging the power of Generative AI~\cite{64,65,66,67,68}, our method virtually creates difficult cases of realistic training data, based on synthetic fish planted into real-world underwater habitats, with additional color and shape transformations that make the resulting images suitable for a highly challenging training task. Our approach works in 2 unsupervised stages. In the first step a segmentation model is trained on images obtained by placing 2D views of synthetic 3D fish in real-world empty habitats. We perform several shape and appearance virtual fish transformations, before placing them into the habitat images: {\bf 1)} random affine transformation~\cite{74}, for varying virtual fish size, orientation and placement; {\bf 2)} Thin Plate Spline shape warping~\cite{76}, to simulate fish undulatory swimming style and {\bf 3)} color Histogram Matching~\cite{75} between fish and empty background, to make its appearance look similar to the underwater habitats, thus increasing the level of realism and difficulty, by not allowing synthetic colors artificially distinguish it from its habitat. After obtaining a relatively powerful Step 1 model, we proceed to Stage 2 where we fine-tune it on images generated by placing synthetic fish in all available training data, which can contain real fish this time. We apply the same fish transformations as before, with the single difference that synthetic fish match now the color histogram of high confidence positive pixels from Stage 1 model's soft outputs, in order to place virtual fish into training images and pseudo-labels which have a similar appearance to real fish from the data. Thus we obtain our final unsupervised Step 2 model, by fine tuning the initial Step 1 model on Stage 2 data. Note that the two stages of our approach are completely unsupervised (without any human intervention), both using for training virtual automatically generated fish, whose appearance matches the background habitat (Stage 1) and real fish from the images (Stage 2) - see Figure~\ref{fig:add}. 


We evaluate our approach on the well-known DeepFish dataset and obtain a performance close to a fully-supervised SoTA segmentation model. 
We also show that our Stage 2 fine-tuning procedure can improve the base supervised model by a significant margin, if we combine our unsupervised virtual annotations with real fish human labels. Additionally, we test the effectiveness of our approach on a novel underwater fish segmentation dataset, DeepSalmon, introduced in this paper and presented in Section \ref{sec:DeepSalmon}.

{\bf Related work on virtual fish data generation and fish augmentation: } Generating synthetic annotated underwater fish data was proposed in a few previous works in the literature~\cite{69,70,71,72}, but, unlike ours, none is completely unsupervised. The most similar to ours is the method introduced in~\cite{73}, where instances of fish are cropped from real human-labeled images and pasted onto empty background habitats at random positions, with random orientations and sizes for training a species classification model. Different from other methods, we place only synthetic fish, for which the segmentation is automatically available, in real underwater images, without relying on any human annotations. Also, our fish augmentation procedure is more complex, involving morphing of shape, motion and appearance that matches the real-world underwater cases scenario.

The majority of existing underwater fish augmentations perform standard data transformations such as flipping, rotation, cropping and random noise~\cite{83, 84, 85, 86}, without taking into account fish characteristics or its relation to the background habitats. Other works~\cite{91} use stable diffusion models to generate diverse synthetic augmented images of particular fish species that can be used for training classification models. However, in order to generate high-quality images the training set should contain enough examples from every species and also their method doesn't provide a way to obtain fish segmentation labels, in contrast to our procedural generative approach, where synthetic images and their corresponding segmentations can be directly obtained, with no additional expenses in terms of human annotation efforts or training costs. Moreover, we generate realistic synthetic images by integrating virtual fish in real underwater background habitats, obtaining models that are more robust to real-world data scenarios, which is difficult to achieve using images that are created purely in a synthetic way.

There are also works in the literature that address the problem of highly-degraded underwater data by enhancing image quality using GANs or Diffusion models~\cite{87,88,89,90}. While these approaches are important for processing and understanding underwater images, they don't tackle the problem of limited existing labeled data and they should rely on human annotations for training fish segmentation models. On the other hand, our method addresses both challenges of low-quality and limited supervision by generating realistic and diverse underwater fish training data, for which the annotations are automatically available.

\section{DeepSalmon Dataset}
\label{sec:DeepSalmon}

There are only few published underwater fish datasets in the literature, such as DeepFish \cite{19} and Seagrass \cite{20}, for fish-only and YouTube VOS \cite{21}, with $94$ object categories, including fish. 
To address the limited annotated fish data, we introduce DeepSalmon, a relatively large fish dataset of 30 GB (see Figure \ref{fig:unsup_both} and \ref{fig:results}), with $12$ difficult videos (at $25$-fps) of \emph{Salmon Salar} species in two built-in systems: a control tank and an ozone tank. Most of the fish in our dataset are hard to detect, even by human eye, due to the poor visibility, delusive appearance of the environment and large number of fish that appear and occlude each other. We provide annotations at both semantic and instance levels for $200$ video frames. Due to the task difficulty, it takes about $30-40$ mins to annotate a single frame. Note that the limited underwater optical view makes it impossible to effectively use label propagation methods \cite{30, 31} for automatic annotation. 

Compared to the existing published fish data, our DeepSalmon dataset has a distinctive appearance, containing underwater scenarios with widely varying fish numbers, sizes and body poses. By testing our Automatic Labelling method on two, very different datasets (DeepFish and DeepSalmon, see Figure \ref{fig:unsup_both} and \ref{fig:results}), we have a good opportunity to analyze the performance of our approach in the real world. Note that DeepFish dataset consists of a large number of videos (40k frames)
collected from $20$ different habitats in remote coastal marine environments
of tropical Australia. It has $3.2k$ point level and $620$ semantic segmentation annotations. In contrast to our dataset, DeepFish has significantly fewer individual fish in the videos, of smaller sizes. 

\section{Automatic Virtual Labelling for Unsupervised Fish Segmentation}

Our proposed method works as follows: having available empty habitats images (without any real fish),  
we place artificially created fish, after performing transformations that change their shape and appearance, to better integrate them into the underwater backgrounds, while also better capturing their undulatory swimming style. Since everything is known about the virtual fish, their ground truth segmentations are automatically available as well. As previously mentioned, in Stage 1 we create synthetic fish with color distributions that are similar to the empty habitats, by matching color histograms between virtual fish and underwater backgrounds, while in Stage 2 of fine-tuning we generate fish with color distributions that are similar to those of real fish from the images. We obtain true fish histograms by collecting the colors of high confidence positive pixels from Stage 1 model soft outputs. Note that without the initial Step 1 model, we would not be able to produce such good quality soft masks to be able to obtain color distributions of real fish from the data.
Below we present in technical details the three types of shape and appearance transformations applied on synthetic fish before placing them into the underwater images:

{\bf 1) Affine transformations} are typically used to correct geometric distortions or deformations that occur with non-ideal camera angles~\cite{74}. In our context we use them to vary synthetic fish size, orientation and placement in the underwater images. The particular affine transformations we applied in our approach are rotation, scaling and translation:
\begin{equation}
\mathcal{F'} = T(t_x, t_y) \cdot S(s_x, s_y) \cdot R(\alpha) \cdot \mathcal{F},
\end{equation}
where $\alpha$ is rotation angle, $s_x, s_y$ are scale factors and $t_x, t_y$ specifies the location in the generated image where fish will be placed.

{\bf 2) Thin Plate Spline (TPS)} transformation~\cite{76} is an image warping algorithm, with applications in image editing~\cite{77}, texture mapping~\cite{78} and image morphing~\cite{79}, where, given a set of $N$ pixels positions from an image $(x_i, y_i)$, named control points, with corresponding displacements $(\Delta x_i, \Delta y_i)$, the idea is to find a function $f : (x,y) \to (x', y')$ that maps pixels from the initial image to pixels from the deformed image so that the warped control points $(x_i', y_i')$ closely match its expected targets $(x_i + \Delta x_i, y_i + \Delta y_i)$, while the surrounding points are deformed as smoothly as possible.

This is done with the help of 2 warping functions $f = (f_{x'}$, $f_{y'})$, that output the deformed position on each corresponding axis, given the initial pixel positions. These 2 functions have the following form:
\begin{equation} 
f_{x'}(x, y) = a_1 + a_x x + a_y y + \sum_{i=1}^N w_i \ \mathcal{U}(\lVert (x_i, y_i) - (x, y) \rVert),
\end{equation}
where $(a_1, a_x, a_y)$ coefficients represents the linear plane that best approximates the warping targets of all control points and $w_i$ terms denotes the weights of the kernels surrounding each control point, where as kernel function is used thin plate spline kernel: $\mathcal{U}(r) = r^2 \log r$. In this way, by passing the distance between a certain point and a control point to this kernel function, the closer the position to a control point is, the higher its displacement corresponding to that point will be. For finding the coefficients of thin plate spline warping function, the following linear system is solved (for $f_{x'}$): 
\begin{equation}
L = \left[ \begin{array}{c|c} 
  K & P \\ 
  \hline 
  P^T & O 
\end{array}\right] \cdot
\left[ \begin{array}{c} 
  w  \\  
  a 
\end{array}\right] = 
\left[ \begin{array}{c} 
  x'  \\  
  0 
\end{array}\right],
\end{equation}
where $K_{ij} = \mathcal{U}(\lVert (x_i, y_i) - (x_j, y_j) \rVert)$, the $i$-th row of $P$ is $(1, x_i, y_i)$ and $x' = (x_0', x_1', \ldots, x_N')$. When solving the system for $f_{y'}$, $x'$ will be replaced with $y' = (y_0', y_1', \ldots, y_N')$.

\begin{figure}
\begin{center}
   \includegraphics[width=0.9\linewidth ]{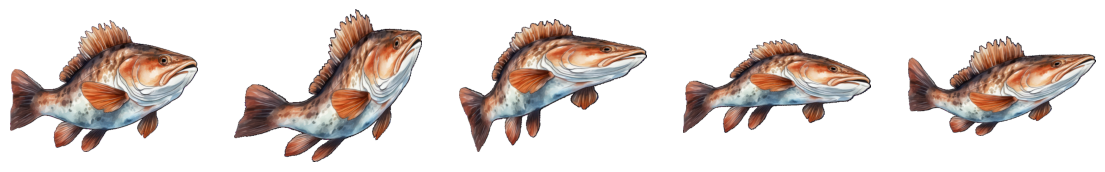}
\hspace{5.2em} \textbf{a)} \hspace{5.2em} \textbf{b)}  \hspace{5.2em}   \textbf{c)} \hspace{5.2em}   \textbf{d)} \hspace{5.2em}   \textbf{e)}
\end{center}
   \caption{Thin plate spline image warping transformation examples \textbf{b) - e)}, when using random pixel positions displacements for a synthetic fish input image \textbf{(a)}. Number of control points is $N=3$.}
\label{fig:tps}
\end{figure}

The local smoothness property of thin plate spline transformation makes it a perfect data augmentation to simulate underwater fish undulatory swimming style (Figure~\ref{fig:tps}). For that reason, we use this image warping method to diversify in a natural way synthetic fish motion and shape in the generated images: $\mathcal{F'}(x,y) = \mathcal{F}(f(x,y))$.

{\bf 3) Histogram Matching}~\cite{75} is a data transformation technique that implies changing pixels distribution for a source image such that its color histogram matches the histogram of a reference image on every channel. This is done by matching the cumulative density functions of the 2 images:
\begin{equation}
\mathcal{F'}(x,y)= G^{-1}(H(\mathcal{F}(x,y))),
\end{equation}
where $H$ and $G$ are CDFs for source and reference images and $G^{-1}$ is obtained by linear interpolation.

Unlike for the previous 2 data augmentations, we perform histogram matching in a different way for the 2 steps of our approach (Figure~\ref{fig:add}). In the first Stage we match the color between synthetic fish and random pixels sampled from the background patch where fish will be placed. In this way we create realistic underwater scenarios, while increasing the difficulty of the Stage 1 training, forcing the unsupervised model to focus on shape rather than appearance, with
shape being realistically modeled by TPS warping applied to synthetic fish.

Then, at the second stage of our method, we sample points from the set of most probable positive pixels in the image, extracted from the pseudo-labels generated by Step 1 model, and we use their color as reference when performing histogram matching. In this manner, we place in the training images synthetic fish that have a similar appearance to the real ones existing in the data.
In order to have a much better augmentation diversity, we sample in both stages of our approach random pixels from the selected area in the image, when creating the reference color histogram. In algorithm \ref{alg:algorithm1} we present the pseudocode for Stage 1 of our method, the only differences for the second step being the appearance of real fish in the training data and the strategy of matching true fish color histograms, as previously explained.

\begin{algorithm}[t]
  \caption{Realistic Shape and Appearance Labeled Data Generation for Unsupervised Fish Segmentation - Stage 1}\label{alg:algorithm1}
  \begin{algorithmic}[1]
    \renewcommand{\algorithmicrequire}{\textbf{Input:}}
	\renewcommand{\algorithmicensure}{\textbf{Output:}}
    \algnewcommand\INPUT{\item[\algorithmicinput]}
    \Require
      \Statex Set of images with empty background habitats \textbf{M}
      \Statex Synthetic fish images \textbf{F} and their corresponding segmentation labels \textbf{L}
      \Statex Data augmentations hyperparameters \textbf{h} and  distribution for the number of added fish \textbf{d}
    \Ensure
      \Statex Stage 1 trained fish segmentation \textbf{Model}
    \vspace{3pt}
        \hrule
    \vspace{3pt}
    \For{every training epoch}
        \For{each background image $\textbf{M}_i$}  
            \State Set $\textbf{I}_i = \textbf{M}_i$ and $\textbf{T}_i = \mathbf{0}$
            \State Get number of fish to be added in the image $\textbf{N} \sim d$
            \For{each fish $\textbf{F}_j$ sampled from $\textbf{F}$, $ j = 1$ to $\textbf{N}$}
            \State Apply affine transformation and TPS warping on $\textbf{F}_j$ and $\textbf{L}_j$ according to 
 \textbf{h} and extract from $\textbf{M}_i$ the background patch \textbf{P} where fish will be placed
            \State Match histograms between $\textbf{F}_j$ and random pixels from \textbf{P}
            \State Paste fish $\textbf{F}_j$ to $\textbf{I}_i$ and its segmentation mask $\textbf{L}_j$ to $\textbf{T}_i$
            \EndFor
            \State Store training example $(\textbf{I}_i, \textbf{T}_i)$
        \EndFor
        \State Train \textbf{Model} on training set (\textbf{I}, \textbf{T})
    \EndFor
  \end{algorithmic}
\end{algorithm}

\begin{figure}
\begin{center}
   \includegraphics[width=0.98\linewidth]{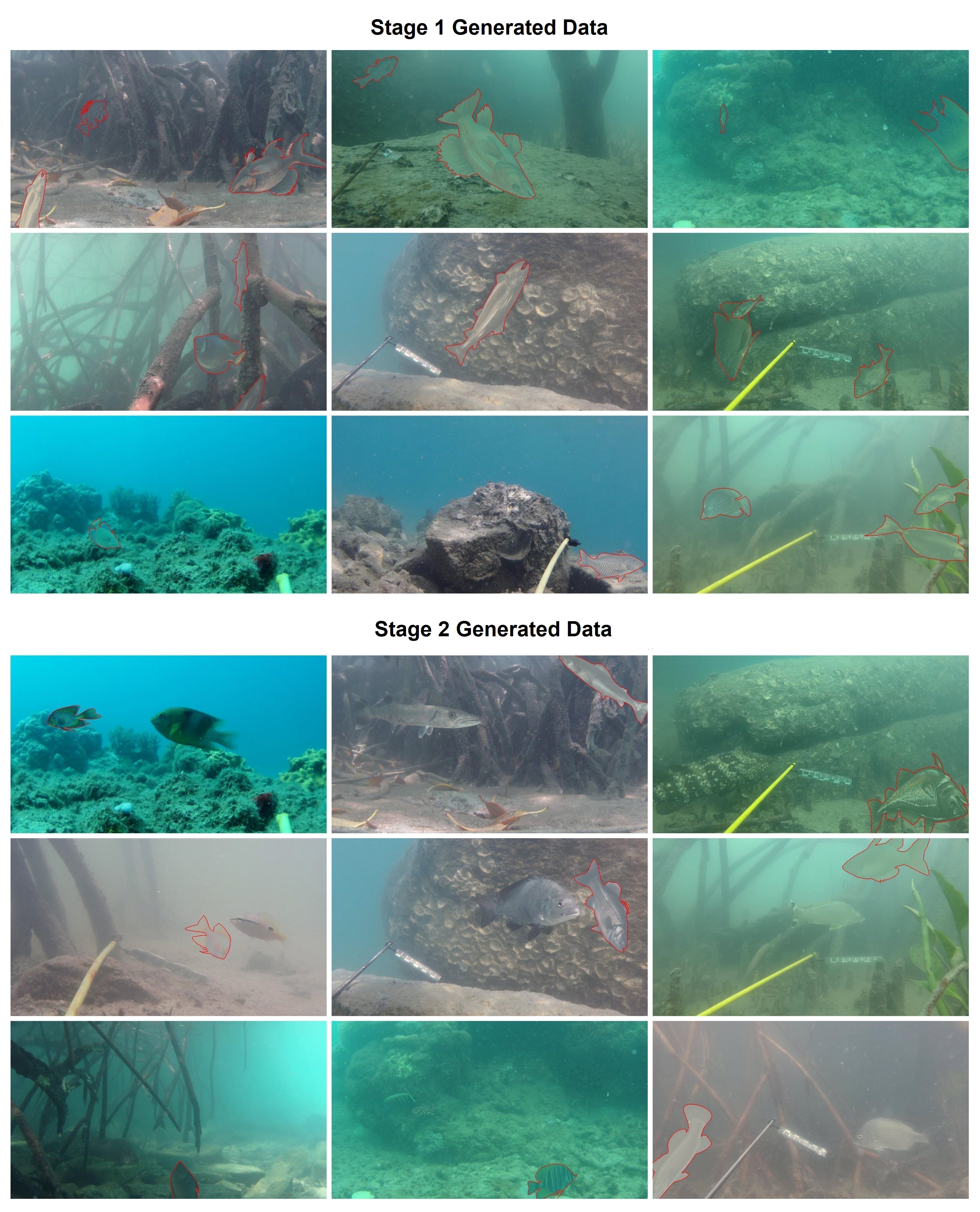}
\end{center}
   \caption{Examples of \textcolor{purple}{synthetic fish} placed into images from DeepFish dataset at both steps of our approach. Note how the generated images become more and more realistic with every stage of our method due to the fish augmentations performed.}
\label{fig:long}
\label{fig:addd}
\end{figure}

\begin{figure}
\begin{center}
   \includegraphics[width=0.99\linewidth]{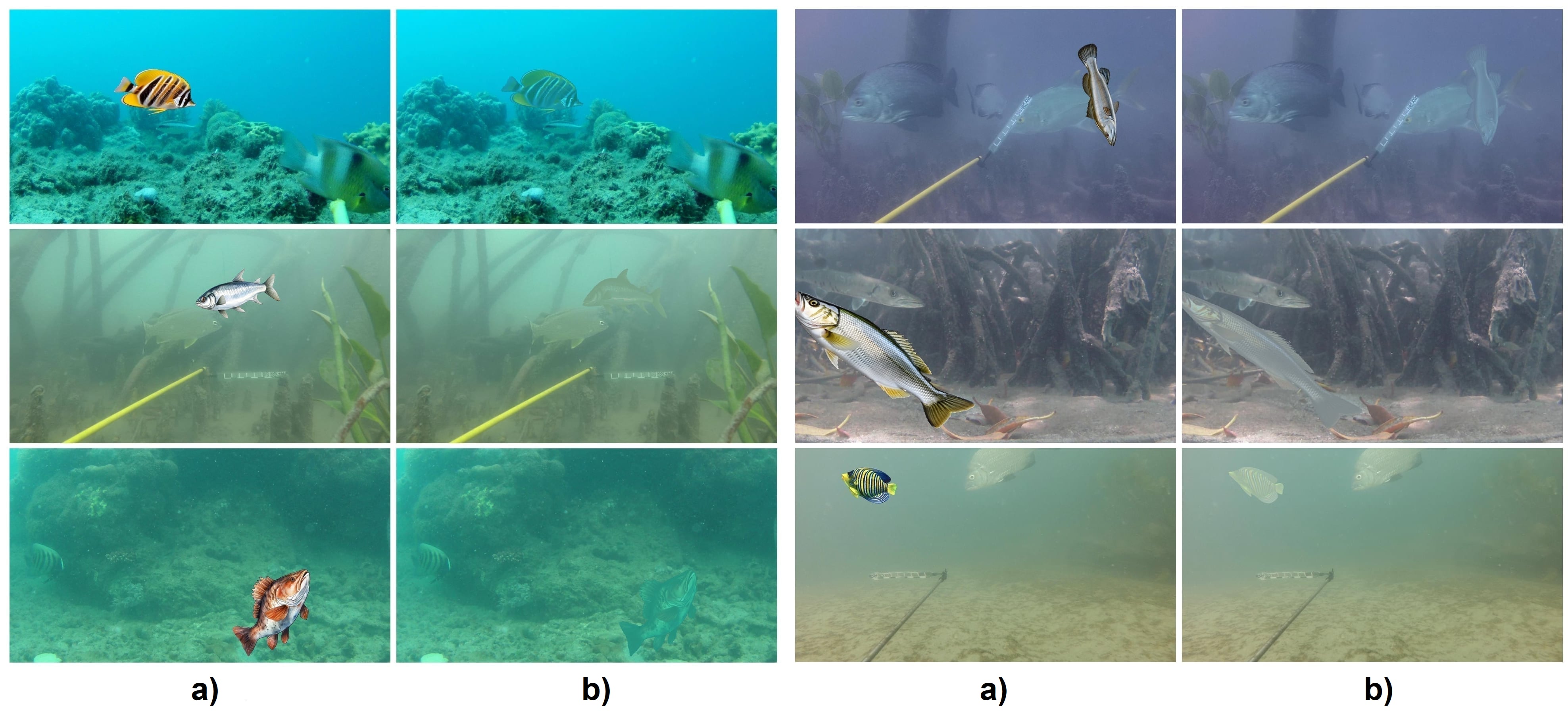}
\end{center}
   \caption{Stage 2 generated data for DeepFish dataset when applying \textbf{(b)} or not \textbf{(a)} Thin-Plate-Spline and color histogram matching virtual fish transformations. Note the unrealistic appearance of the generated images when these augmentations are not used.}
\label{fig:long}
\label{fig:compare}
\end{figure}

We illustrate in Figure~\ref{fig:addd} examples of images generated in the 2 steps of our approach on DeepFish dataset. Note how by combining the difficult challenge of camouflage (by matching background color) and naturalistic undulatory shape morphing (using TPS warping) at Stage 1 with good knowledge of true fish appearance (by matching real fish color) at Stage 2, we obtain a completely
unsupervised method in which each stage brings the generated images closer to the real world underwater fish data. In order to highlight this property of our approach, we present in Figure~\ref{fig:compare} how the data generated at Stage 2 would look if we don't apply Thin-Plate-Spline and color histogram matching fish transformations. Note that if our proposed data augmentations are not used, synthetic fish appearance will look unnatural and out of context in the generated images.

\section{Experiments}

We validate our Realistic Shape and Appearance Data Generation method by testing it on DeepFish and our newly introduced DeepSalmon datasets, showing how our approach produce qualitative segmentation results, using no human annotation, on 2 completely different underwater datasets, considering fish and habitat appearance, highlighting its robustness and efficiency in the real world.

\subsection{Experimental Setup}
\begin{figure}
\begin{center}
   \includegraphics[width=0.9\linewidth, ]{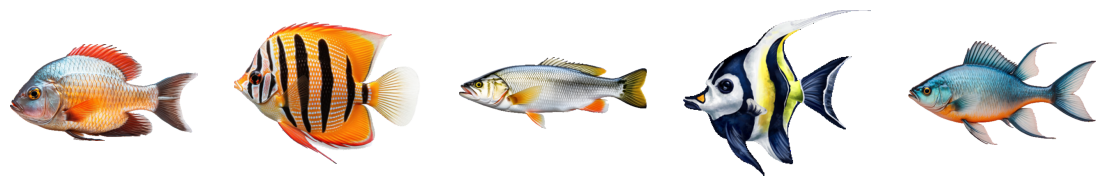}
\end{center}
   \caption{Examples of synthetic fish placed into real underwater fish images in our Realistic Shape and Appearance Data Generation method.}
\label{fig:fish}
\end{figure}

In order to show the effectiveness of our approach, we used in all our experiments a fixed set of only $100$ synthetic fish images with transparent backgrounds (examples shown in Figure~\ref{fig:fish}), which are randomly selected from Vecteezy stock photos public website (\textit{https://www.vecteezy.com}), using no prior knowledge about the appearance of real fish from the 2 evaluated datasets. As image segmentation model we used the State of The Art SegFormer, ver. b4~\cite{11} and the 2 datasets have a train-validation-test split of $80\%-10\%-10\%$ for both DeepFish and DeepSalmon. At first stage of our approach we train the segmentation model $50$ epochs with a learning rate of $10^{-4}$ by adding synthetic fish in empty underwater habitat images, extracted from the 2 datasets. Then, at the second step, we fine-tune Stage 1 model on the entire training set using a learning rate of $7 \cdot 10^{-5}$ multiple iterations, until no improvement is obtained from one iteration to the next one, where a iteration consists of 5 epochs of training. The pseudo-labels, which are computed for the best Dice score (F-measure) over Precision-Recall curve, are updated after each iteration of fine-tuning.

The hyperparameters used in our experiments for transforming synthetic fish before placing them into the underwater images  are the following: \textbf{1)} rotation and translation factors cover the range of all possible values such as at least half of the fish can be seen in the images and the ratio between synthetic fish and underwater images largest dimensions is sampled from the interval $[1/10, 1/3]$ for DeepFish and $[1/5, 2/3]$ for DeepSalmon, since in our dataset appear fish with a larger size; \textbf{2)} for TPS warping we used $N=3$ control points and the displacements are sampled uniformly within range $[-0.2 x_i, 0.2 x_i]$ where $x_1, x_2$ represents the 2 dimensions of the synthetic fish image; \textbf{3)} when performing color histogram matching, we consider at Stage 2 the positive pixels with a confidence over $0.8$ from the pseudo-labels only if exists at least $1\%$ such pixels in the image (otherwise, since most likely there are no real fish in the image, we choose not to place any synthetic fish as well). For both stages of our approach we sample $10\%$ random pixels from the selected area in the image, when creating the reference
histogram color, in order to obtain a better augmentation diversity. \textbf{4)} The distributions for the number of synthetic fish to be added in the images for the 2 steps of our method are: at first stage  $P(d_1 = i) = 1/N$ for all $i = 1$ to $N$, where $N$, the maximum number of fish, is $3$ in case of DeepFish and $N = 6$ for our dataset; at the second stage $d_2 = \begin{pmatrix}
 0 & 1 \\
 0.2 & 0.8
\end{pmatrix}$ and $d_2 = \begin{pmatrix}
 0 & 1 & 2 \\
 0.2 & 0.4 & 0.4
\end{pmatrix}$ for DeepFish and gDeepSalmon, respectively. Using these probability distributions, we take into account the higher number of fish appearing in our dataset and we let the model see at some epochs from our fine-tuning stage the initial data, without synthetic fish added. Note that all hyperparameters selections were made without using any statistics from neither of the 2 datasets.

\subsection{Experimental Results}
In Table~\ref{table:tab1} we present the results, in terms of Dice and IoU metric scores, obtained by the 2 steps of our unsupervised Automated Virtual Labelling method on DeepFish and DeepSalmon, 2 dataset with different fish and habitat appearance. Note how, by performing realistic fish transformations that simulates with high fidelity real underwater fish scenarios, our Stage 1 segmentation model have a relatively good performance on the 2 evaluated fish datasets, which is further significantly boosted by our second step procedure ($+ 22$ IoU score in case of both datasets), obtaining a final performance close to the fully-supervised case. We compared Stage 2 of our approach with the classical semi-supervised strategy, where our Step 1 model is fine-tuned only on pseudo-labels, without adding synthetic fish in the training images. Although this baseline method also produces an important improvement over the initial model (which is caused by model adaptation from synthetic to real fish), the second step of our approach offers an additional $+8$ IoU score.

\begin{table}[!t]
\begin{center}
\begin{tabular}{@{\extracolsep{0.3pt}}lccccccccc@{}}
\toprule
& \multicolumn{4}{c}{DeepFish} & \multicolumn{4}{c}{DeepSalmon} \\
\cmidrule(lr){2-5} \cmidrule(r){6-9}
& \multicolumn{2}{c}{Unsupervised} & \multicolumn{2}{c}{Supervised} & \multicolumn{2}{c}{Unsupervised} & \multicolumn{2}{c}{Supervised} \\
\cmidrule(lr){2-3} \cmidrule(lr){4-5} \cmidrule(lr){6-7} \cmidrule(r){8-9}
Models & Dice & IoU & Dice & IoU & Dice & IoU & Dice & IoU \\
\midrule
Our Stage 1 & 78.53 & 64.64 & 95.44 & 91.27 & 77.11 & 62.74 & 96.23 & 92.73 \\
Baseline Fine-Tuning & 87.81 & 78.26 & 95.65 & 91.66 & 86.41 & 76.07 & 96.17 & 92.62 \\
Our Stage 2 & \textbf{92.54} & \textbf{86.11} & \textbf{96.20} & \textbf{92.67} & \textbf{91.94} & \textbf{85.08} & \textbf{96.42} & \textbf{93.09} \\
\bottomrule
\end{tabular}
\end{center}
\caption{Metric scores for the two stages of our approach and the baseline semi-supervised strategy for fine-tuning Stage 1 model on DeepFish and DeepSalmon datasets in both unsupervised (when training Step 1 model from scratch) and supervised (when using the fully-supervised model as Step 1 model) settings.}
\label{table:tab1}
\end{table}

\begin{table}[!t]
\begin{center}
\begin{tabular}{@{\extracolsep{0.3pt}}lccccccccc@{}}
\toprule
& \multicolumn{4}{c}{DeepFish} & \multicolumn{4}{c}{DeepSalmon} \\
\cmidrule(lr){2-5} \cmidrule(r){6-9}
& \multicolumn{2}{c}{Unsupervised} & \multicolumn{2}{c}{Supervised} & \multicolumn{2}{c}{Unsupervised} & \multicolumn{2}{c}{Supervised} \\
\cmidrule(lr){2-3} \cmidrule(lr){4-5} \cmidrule(lr){6-7} \cmidrule(r){8-9}
Models & Dice & IoU & Dice & IoU & Dice & IoU & Dice & IoU \\
\midrule
Our Stage 2 & \textbf{92.54} & \textbf{86.11} & \textbf{96.20} & \textbf{92.67} & \textbf{91.94} & \textbf{85.08} & \textbf{96.42} & \textbf{93.09} \\
w/ all positives HM & 91.92 & 85.04 & 96.14 & 92.56 & 91.78 & 84.80 & 96.40 & 93.05 \\
w/ Stage 1 HM & 89.34 & 80.73 &  95.58 & 91.53 & 90.24 & 82.21 & 96.26 & 92.78 \\
w/o HM & 89.06 & 80.27 & 95.61 & 91.58 & 90.09 & 81.96 & 96.24 & 92.75 \\
w/o TPS & 90.52 & 82.68 & 96.11 & 92.51 & 90.42 & 82.51 & 96.32 & 92.90 \\
w/o HM and TPS & 88.35 & 79.13 &  95.53 & 91.44 & 88.11 & 78.74 & 96.25 & 92.77 \\
\bottomrule
\end{tabular}
\end{center}
\caption{Ablation study for the second stage of our approach. Note the importance of both proposed fish transformations: color histogram matching (HM) and thin plate spline (TPS) shape warping.}
\label{table:tab2}
\end{table}
We also show in Table~\ref{table:tab1} how the second stage of our method can improve with a good margin the performance of the fully-supervised model as well, when using this model for generating pseudo-labels in our approach, instead of the unsupervised Stage 1 soft outputs. The results show how our strategy of adding realistic synthetic fish with various shapes and a similar appearance to the real fish from the data at fine-tuning step helps the supervised model to considerably improve metric scores (Dice error is reduced by $16 \%$ on DeepFish). Moreover, since this model is initially trained on real fish images in this case, the boost over baseline semi-supervised fine-tuning is more significant for our approach.

\begin{figure}
\begin{center}
   \includegraphics[width=0.98\linewidth]{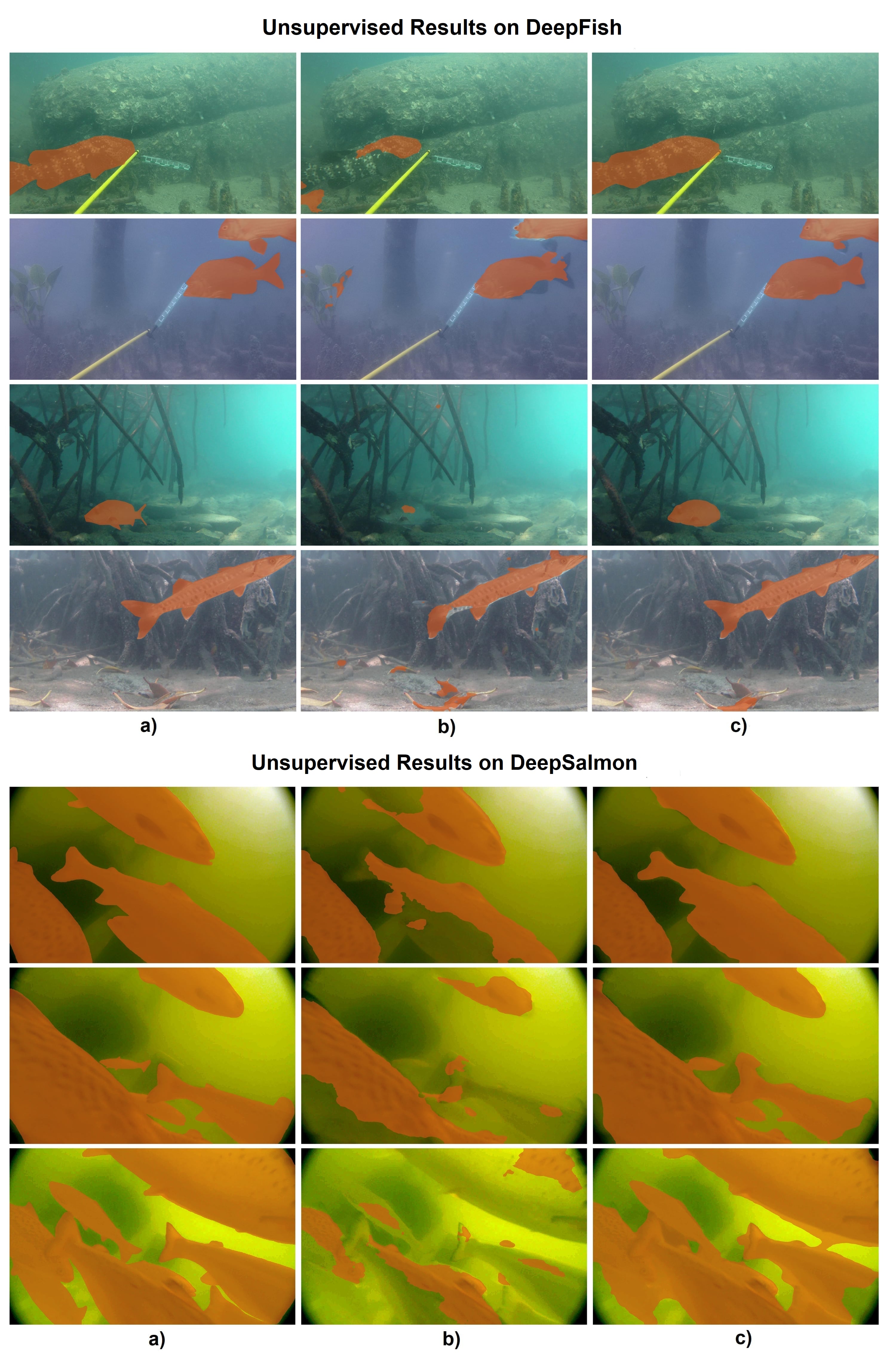}
\end{center}
   \caption{Qualitative segmentation results on DeepFish and DeepSalmon. On each row, images show, in order, \textbf{a)} input frame and its corresponding ground truth, \textbf{b)} Stage 1 model prediction and \textbf{c)} final Stage 2 prediction for our unsupervised method.}
\label{fig:unsup_both}
\end{figure}

Figure~\ref{fig:unsup_both} shows qualitative segmentation results produced by our models on DeepFish and our DeepSalmon datasets in the unsupervised setting. Note how our fine-tuning strategy helps the first stage model to detect previously unseen fish of different sizes and shapes and to improve the segmentation of the already discovered fish from the images. 

\begin{figure}
\begin{center}
   \includegraphics[width=0.98\linewidth]{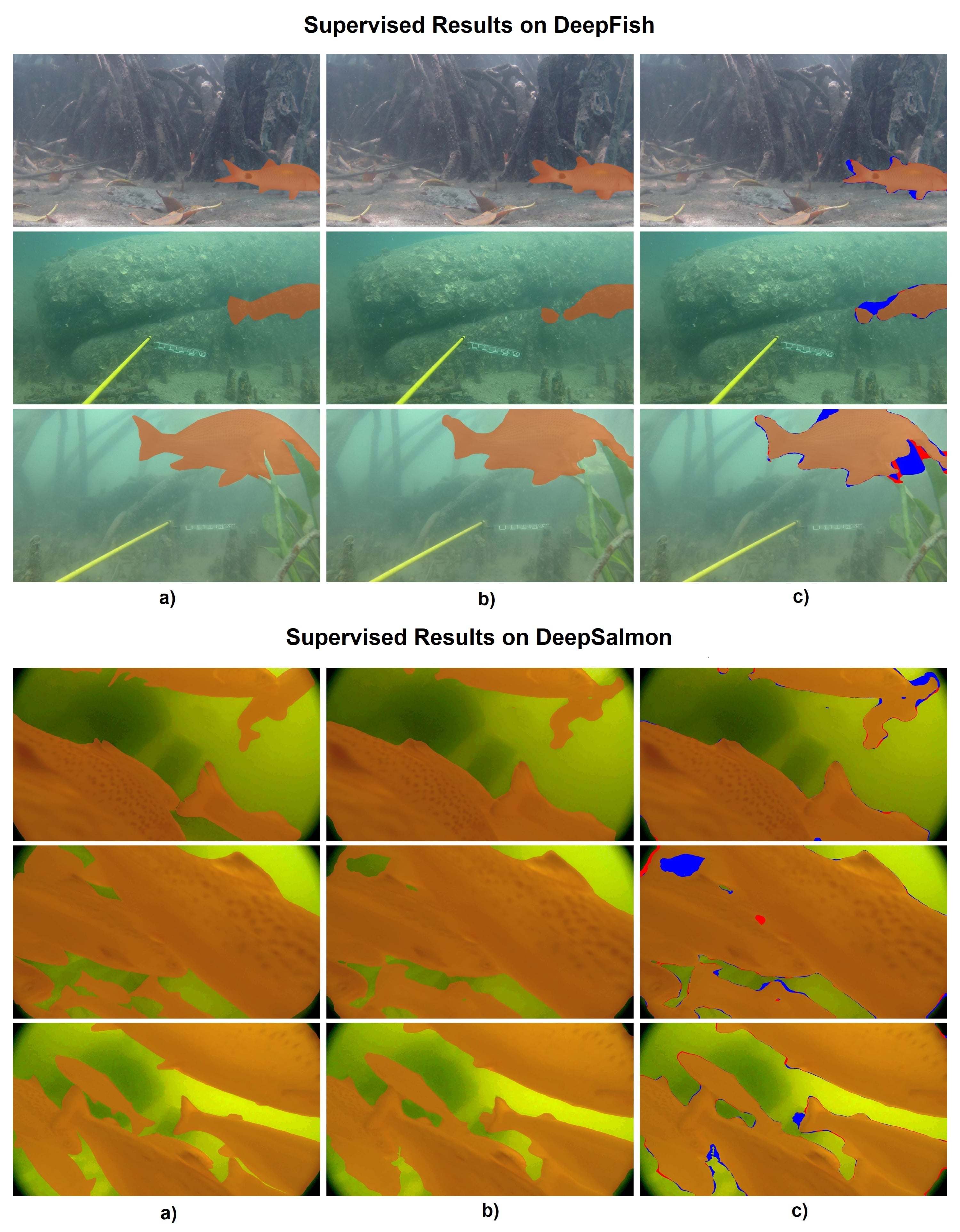}
\end{center}
   \caption{Qualitative segmentation results on DeepFish and DeepSalmon datasets in the supervised setting. On each row, the images illustrate, in order, \textbf{a)} input frame and its ground truth, \textbf{b)} supervised model prediction and \textbf{c)} our Stage 2 model output. Pixel predictions where our fine-tuned model differ from the initial supervised model are shown in \textcolor{blue}{blue} (where we are correct) or \textcolor{red}{red} (where we are wrong).}
\label{fig:long}
\label{fig:results}
\end{figure}

Moreover, we illustrate in Figure~\ref{fig:results} additional results produced by our approach on both datasets in the supervised setting. Pixel predictions where our Stage 2 model differ from the initial fully-supervised model are shown in blue (where we are correct) and red (where we are wrong). The results show how our method helps the supervised model to detect formerly unseen fish parts and to further refine fish segmentation masks, which were already qualitative.

In Table~\ref{table:tab2} we present an ablation study for the second stage of our approach. The results show the importance of both TPS and color histogram matching fish transformations, the performance dropping significantly when we don't apply one of these two proposed augmentations and further more if we don't use any of them (when performing only the standard affine transformations). Moreover, our analysis demonstrates the efficiency of matching color distribution of most probable positive pixels at Stage 2, over other strategies such as using all positive pixels from pseudo-labels or matching background histograms, as in Stage 1.

\begin{table}[!h]
\begin{center}
\begin{tabular}{lccccc}
\toprule
 & {Our Stage 2} & {~\cite{82}} & \text{    LCFCN + PM}~\cite{81} & \text{       A-LCFCN}~\cite{81} & \text{     LCFCN}~\cite{80} \\
\midrule
IoU & \textbf{86.11} & \text{-} & 73.00 & 71.30 & 68.40 \\
AP &  \textbf{87.34}  & 75.00 & \text{-} & \text{-} & \text{-} \\

\bottomrule
\end{tabular}
\end{center}
\caption{A comparison between our approach and other existing unsupervised~\cite{82} and weakly supervised~\cite{80, 81}  underwater fish segmentation methods on DeepFish dataset. }
\label{table:tab3}
\end{table}

Lastly, we compared in Table~\ref{table:tab3} the performance of our unsupervised model with other few existing fish segmentation approaches in the literature that report results on DeepFish. Without using any human annotations, our algorithm obtains a much better IoU score than weakly-supervised methods introduced in~\cite{80, 81}, where point-level annotations are used for generating pseudo-labels. Also, our approach that can be trained directly on still images, since it doesn't rely on temporal variations from videos, produce a better Average Precision score than unsupervised solution proposed in~\cite{82}, where background subtraction and optical flow are used for creating pseudo-labels. Note that in case of some underwater videos, like those existing in our dataset, the limited optical view makes these temporal-based techniques impossible to use effectively, while our method can produce qualitative results in this case as well (see Table~\ref{table:tab1}).

\section{Conclusions}
We introduced in this paper a highly effective approach for generating automatic labeled underwater fish data by placing virtual fish in real underwater habitats, which works in 2 completely unsupervised stages that make the virtual images increasingly closer to the real-world scenarios. At Stage 1 we introduce synthetic fish in real empty underwater habitats that match the background color distribution, which makes fish look realistic, while giving a challenging training task for the initial segmentation model. Fish camouflage forces Step 1 model to learn especially shape properties of the fish, which, combined with the proposed Thin Plate Spline form warping, results in a strong and robust Stage 1 model. This is followed by Stage 2 fine tuning, where the virtual fish introduced in the images have now a similar appearance to real fish from the data, by matching the color distribution of most probable positive pixels from Step 1 outputs.
The two stages of our unsupervised approach bring together the performance to a level that is very close to a fully-supervised SoTA model. While our experiments focus on the difficult task of underwater fish segmentation, on two very different
datasets (one established and one introduced by us), our approach is in fact relatively general, being applicable to essentially any type of non-rigid object class.

\section*{Acknowledgements}
This work is supported by projects “Unleashing the Sustainable Value Creation Potential of Offshore Ocean” (Grant No. 328724) and “The balancing act: Biologically driven rapid-response automation of production conditions in recirculating aquaculture systems (RAS)” (Grant No. 320717), co-funded by the Research Council of Norway, and EU Horizon project ELIAS (Grant No. 101120237).

%
%
\bibliographystyle{splncs04}
\bibliography{main}
\end{document}